%% file: main_preprint.tex
\documentclass{article}

\usepackage{preprint}

\usepackage[utf8]{inputenc} 
\usepackage[T1]{fontenc}    
\usepackage{hyperref}       
\usepackage{url}            
\usepackage{booktabs}       
\usepackage{amsfonts}       
\usepackage{nicefrac}       
\usepackage{microtype}      
\usepackage{lipsum}

\usepackage[utf8]{inputenc} 
\usepackage[T1]{fontenc}    
\usepackage{hyperref}       
\usepackage{url}            
\usepackage{booktabs}       
\usepackage{amsfonts}       
\usepackage{nicefrac}       
\usepackage{microtype}      
\usepackage{soul}
\usepackage{booktabs} 
\usepackage{amsmath}
\usepackage{mathtools}
\usepackage[toc,page]{appendix}
\usepackage{algcompatible}
\usepackage{algorithm}
\usepackage{varwidth}
\usepackage{longtable}
\usepackage{subcaption}
\usepackage{comment}
\usepackage{enumerate}
\usepackage{multicol}        
\usepackage[bottom]{footmisc}
\usepackage{newtxtext}       %
\usepackage{newtxmath}       
\usepackage{lscape}
\usepackage{array}
\usepackage[toc,page]{appendix}
\algnewcommand\algorithmicforeach{\textbf{for each}}
\algdef{S}[FOR]{ForEach}[1]{\algorithmicforeach\ #1\ \algorithmicdo}
\usepackage{xcolor}
\usepackage[noend]{algpseudocode}
\DeclareCaptionType{mycapequ}[Equation][]
\algnewcommand{\LineComment}[1]{\State \(\triangleright\) {\footnotesize #1}}
\algnewcommand{\algorithmicand}{\textbf{ and }}
\algnewcommand{\algorithmicor}{\textbf{ or }}
\algnewcommand{\OR}{\algorithmicor}
\algnewcommand{\AND}{\algorithmicand}
\algnewcommand{\var}{\texttt}
\algnewcommand\algorithmicforeach{\textbf{for each}}
\algdef{S}[FOR]{ForEach}[1]{\algorithmicforeach\ #1\ \algorithmicdo}
\usepackage{hyperref}
\usepackage{soul}

\newcommand{\etal}{{\it et al}}
\newcommand{\eg}{{\textit{e.g.}}}
\newcommand{\ie}{{\textit{i.e.}}}

\definecolor{darkgreen}{rgb}{0,0.3922,0}

\newcommand{\citet}[1]{\citeauthor{#1} \shortcite{#1}} 
\newcommand{\citep}{\cite}

\title{An Empirical Exploration of Deep Recurrent Connections and Memory Cells Using Neuro-Evolution}

\author{
  Travis J. Desell\\
  \texttt{tjdvse@rit.edu} \\
    \And
  AbdElRahman A. ElSaid \\
  \texttt{aelsaid@mail.rit.edu} \\
  \\
    Golisano College of Computing and Information Sciences\\
    Rochester Institute of Technology\\
    Rochester, NY 14623\\
   \And
 Alexander G. Ororbia\\
  \texttt{ago@cs.rit.edu} \\
}

\begin{document}
\maketitle

\input{00-abstract}
\input{01-motivation}
\input{02-related_work}

\input{05-data}
\input{06-results}

\input{07-discussion}

\section{Acknowledgements}
This material is in part supported by the U.S. Department of Energy, Office of Science, Office of Advanced Combustion Systems under Award Number \#FE0031547 and by the Federal Aviation Administration National General Aviation Flight Information Database (NGAFID) award. We also thank Microbeam Technologies, Inc., as well as Mark Dusenbury, James Higgins, Brandon Wild at the University of North Dakota for their help in collecting and preparing the coal-fired power plant and NGAFID data, respectively.

\bibliographystyle{unsrt}  
\bibliography{./references.bib}

\end{document}

%% file: 00-abstract.tex
Neuro-evolution and neural architecture search algorithms have gained increasing interest due to the challenges involved in designing optimal artificial neural networks (ANNs). While these algorithms have been shown to possess the potential to outperform the best human crafted architectures, a less common use of them is as a tool for analysis of ANN structural components and connectivity structures. 
In this work, we focus on this particular use-case to develop a rigorous examination and comparison framework for analyzing recurrent neural networks (RNNs) applied to time series prediction using the novel neuro-evolutionary process known as Evolutionary eXploration of Augmenting Memory Models (EXAMM). Specifically, we use our EXAMM-based analysis to investigate the capabilities of recurrent memory cells and the generalization ability afforded by various complex recurrent connectivity patterns that span one or more steps in time, i.e., deep recurrent connections. 
EXAMM, in this study, was used to train over $10.56$ million RNNs in $5,280$ repeated experiments with varying components. 
While many modern, often hand-crafted RNNs rely on complex memory cells (which have internal recurrent connections that only span a single time step) operating under the assumption that these sufficiently latch information and handle long term dependencies, our results show that networks evolved with deep recurrent connections perform significantly better than those without. More importantly, in some cases, the best performing RNNs consisted of only simple neurons and deep time skip connections, \emph{without any memory cells}. 
These results strongly suggest that utilizing deep time skip connections in RNNs for time series data prediction not only deserves further, dedicated study, but also demonstrate the potential of neuro-evolution as a means to better study, understand, and train effective RNNs. 

%% file: 01-motivation.tex
\section{Introduction}
\label{sec:motivation}

Neural architecture search poses a challenging problem since the possible search space for finding optimal or quasi-optimal architectures is massive. For the case of recurrent neural networks (RNNs), this problem is further confounded by the fact that every node in its architecture can be potentially connected to any other node via a recurrent connection which passes information stored in a vector history to the current time step. Complexity is further increased when one considers that recurrent connections could explicitly connect information from any time step $<t$ in the history of the sequence processed so far to step $t$, improving memory retention through time delays. 
Most modern-day RNNs simplify the recurrent connectivity structure and instead improve retention by utilizing memory cells such as $\Delta$-RNN units~\cite{ororbia2017diff}, gated recurrent units (GRUs)~\cite{chung2014empirical}, long short-term memory cells (LSTMs)~\cite{hochreiter1997long}, minimal gated units (MGUs)~\cite{zhou2016minimal}, and update gate RNN cells (UGRNNs)~\cite{collins2016capacity}. The use of memory cells, as opposed to investigating the use of denser temporal/recurrent connectivity structures, is popular largely under the assumption that, while the recurrent synapses that define a cell only explicitly connect $t-1$ to $t$, their latch-like behavior is sufficient for capturing enough information about the sequence observed so far when making predictions of what will come next. Nonetheless, RNNs still struggle to effectively learn long-term dependencies in temporal data and the quest for the optimal memory cell continues to this day \cite{bengio1993problem,bengio1994learning,hochreiter2001gradient,pascanu2013difficulty,ororbia2017diff,jing2019gated}.

There, however, exists a body of literature that suggests that recurrent connections which skip more than a single time step, which we will coin as \emph{deep recurrent connections}, can play an important role in allowing an RNN to more effectively capture long-term temporal dependencies.
This research dates back to Lin \etal's development of NARX (Nonlinear AutoRegressive eXogenous Model) neural networks with increasing embedded memory orders (EMOs) or time windows~\cite{lin1996learning,lin1998embedded}, which involved adding recurrent connections up to a specified number time skips. Further work went on to show that the order of a NARX network is crucial in determining how well it will perform -- when the EMO of a NARX model matches the order of an unknown target recursive system strong and robust generalization is achieved~\cite{lin1998remember,giles1998past}. Diaconescu later utilized these EMO-based NARX networks to predict chaotic time series data, with best results found in the EMO ranges of $12$ to $30$, which are significantly large time skips~\cite{diaconescu2008use}.
More generally, it has been expressed in classical literature that skip connections can substantially express the computational abilities of artificial neural networks (ANNs) \cite{mcclelland1987parallel}. Yet, modern popular ANNs have only taken advantage of feedforward skip connections \cite{he2016deep}, including RNNs \cite{graves2013generating,srivastava2015highway}, with a few notable exceptions \cite{zilly2017recurrent}.

Findings for RNNs with deep recurrent connections are also not limited to Lin \etal's EMO NARX networks. Chen and Chaudhari developed a segmented-memory recurrent neural network (SMRNN)~\cite{chen2009segmented}, which utilizes a two layer recurrent structure which first passes input symbols to a symbol layer, and then connects the symbol layers to a segmentation layer. This work showed that intervals $10 <= d <= 50$ provided the best results on this data, as a lower $d$ required more computation each iteration (the segmentation was used too frequently) slowing convergence, and at higher values of $d$ it approximated a conventional RNN (that did not use a segmentation layer). The segment interval $d$ operates similarly to a deep recurrent connection; it passes information from past states further forward along the unrolled network.  It was shown that SMRNN outperformed both LSTM and Elman RNNs on the latching problem. ElSaid \etal~later utilized time-windowed LSTM RNNs to predict engine vibration in time series data gathered from aircraft flight data recorders~\cite{elsaid2016using,elsaid2018using}. This work investigated a number of architectures and found that a two-level system with an EMO/time window of order $10$ provided good predictions of engine vibration up to $20$ seconds in the future. This was a challenging problem due to the spiking nature of engine vibration, yet this architecture significantly outperformed time-windowed NARX models, Nonlinear Output Error (NOE), and the Nonlinear Box–Jenkins (NBJ) models.

In this work, we further investigate power of deep recurrent connections in comparison to memory cells by taking a rather unconventional approach to the analysis, using an neuro-evolutionary algorithm we call EXAMM (Evolutionary eXploration of Augmenting Memory Models)~\cite{ororbia2019examm}. Instead of simply testing a few hand-crafted RNNs with and without deep recurrent connections composed of different kinds of memory cells, neuro-evolution was used to select and mix the architectural components as well as decide the depth and density of the connectivity patterns, facilitating an exploration of the expansive, combinatorial search space when accounting for the many different components and dimensions one could explore -- yielding a more rigorous, comprehensive yet automated examination. 
A variety of experiments were performed evolving RNNs consisting of simple neurons or memory cells, e.g., LSTM, GRU, MGU, UGRNN, $\Delta$-RNN cells, as well as exploring the option of using deep recurrent connections or not, of varying degree and instensity.  RNNs were evolved with EXAMM to perform time series data prediction on four real world benchmark problems. In total, $10.56$ million RNNs were trained to collect the results we report in this study.

The findings of our EXAMM-driven experimentation uncovered that networks evolved with deep recurrent connections perform significantly better than those without, and, notably, in some cases, the best performing RNNs consisted of only simple neurons with deep recurrent connections (\ie, no memory cells). 
These results strongly suggest that utilizing deep recurrent connections in RNNs for time series data prediction not only warrants further study, but also demonstrates that neuro-evolution is a potentially powerful tool for studying, understanding, and training effective RNNs. Another salient result from our findings is that, for time series prediction, the relatively new $\Delta$-RNN cell performs better and more reliably than other memory cells.

%% file: 02-related_work.tex
\section{Evolving Recurrent Neural Networks}
\label{sec:related_work}

Neuro-evolution, or the use of artificial evolutionary processes (such as genetic algorithms \cite{goldberg1988genetic}) to automate the design of artificial neural networks (ANNs), has been well applied to feed forward ANNs for tasks involving static inputs, including convolutional variants~\cite{salama2014novel,suganuma2017genetic,sun-arxiv-evocnn-2017,mikkulainen2017codeepneat,stanley2002evolving,stanley2009hypercube}. However, significantly less effort has been put into exploring the evolution of recurrent memory structures that operate with complex sequences of data points. 

Despite the current lack of focus on RNNs, several neuro-evolution methods have been proposed evolving RNN topologies (along with weight values themselves) with NeuroEvolution of Augmenting Topologies (NEAT)~\cite{stanley2002evolving} perhaps being the most well-known. Recent work by Rawal and Miikkulainen investigated an information maximization objective~\cite{rawal2016evolving} strategy for evolving RNNs, which essentially operates similarly to NEAT except with LSTM cells being used instead of simple (traditional) neurons. Research centered around this line of NEAT-based approaches has also explored the use of a tree-based encoding~\cite{rawal-evolving-rnns-2018} to evolve recurrent cellular structures within fixed architectures composed of multiple layers of the evolved cell types. More recently, work by Camero \etal~has shown that a Mean Absolute Error (MAE) random sampling strategy can provide good estimates of RNN performance~\cite{camero2018low}, successfully incorporating it into an LSTM-RNN neuro-evolution strategy~\cite{camero2019specialized}. However, none of this prior work has investigated the evolution deep recurrent connectivity structures nor focused on using a powerful evolutionary strategy such as EXAMM as an empirical analysis tool for RNNs.

With respect to other nature-inspired metaheuristic approaches for evolving RNNs, ant colony optimization (ACO) has also been investigated~\cite{desell2015evolving} as a way to select which connections should be used but only for single time-step Elman RNNs. ACO has also been used to reduce the number of trainable connections in a fixed time-windowed LSTM architecture by half while providing significantly improved prediction of engine vibration~\cite{elsaid2018optimizing}. 

For this study, EXAMM was selected as the RNN analysis algorithm for a number of reasons. First, this procedure progressively grows larger ANNs in a manner similar to NEAT which stands in contrast to current ACO-based approaches, which have been often restricted to operating within a fixed neural topology. 
Furthermore, in contrast to the well-known NEAT, EXAMM utilizes higher order node-level mutation operations, Lamarckian weight initialization (or the reuse of parental weights), and back-propagation of errors (backprop) \cite{rumelhart1988learning} to conduct local search, the combination of which has been shown to speed up both ANN training as well as the overall evolutionary process.  Unlike the work by Rawal and Miikkulainen, EXAMM operates with an easily-extensible suite of memory cells, including LSTM, GRU, MGU, UGRNN, $\Delta$-RNN cells and, more importantly, has the natural ability to evolve deep recurrent connections over large, variable time lags. In prior work it has also been shown to more quickly and reliably evolve RNNs in parallel than training traditional layered RNNs sequentially~\cite{desell-evostar-2019}. For detailed EXAMM implementation details we refer the reader to~\cite{ororbia2019examm}.



%% file: 05-data.tex
\section{Experimental Data}
\label{sec:data}

This experimental study utilized two open-access real-world data sets as benchmark problems for evolving RNNs that can predict four different time series parameters. The first dataset comes from a selection of $10$ flights worth of data taken from the National General Aviation Flight Information Database (NGAFID) and the other comes from data collected from 12 burners of a coal-fired power plant. 
Both datasets are multivariate (with $26$ and $12$ parameters, respectively), non-seasonal, and the parameter recordings are not independent.  Furthermore, the underlying temporal sequences are quite long -- the aviation time series range from $1$ to $3$ hours worth of per-second data while the power plant data consists of $10$ days worth of per-minute readings. To the authors' knowledge, other real world time series data sets of this size and at this scale are not freely available. These datasets are freely provided in the EXAMM github repository\footnote{https://github.com/travisdesell/exact}.

\subsection{Aviation Flight Recorder Data}
\label{sec:avaiation}
Each of the $10$ flight data files last over an hour and consist of per-second data recordings from $26$ parameters, including engine parameters such as engine cylinder head temperatures, gasket temperatures, oil temperature and pressure, and rotations per minute (RPM); flight parameters such as altitude above ground level, indicated air speed, lateral and normal acceleration, pitch, and roll; and environmental parameters such as outside air temperature and wind speed. The data is provided raw and without any normalization applied.

\begin{figure}
    \centering
    \includegraphics[width=0.98\textwidth]{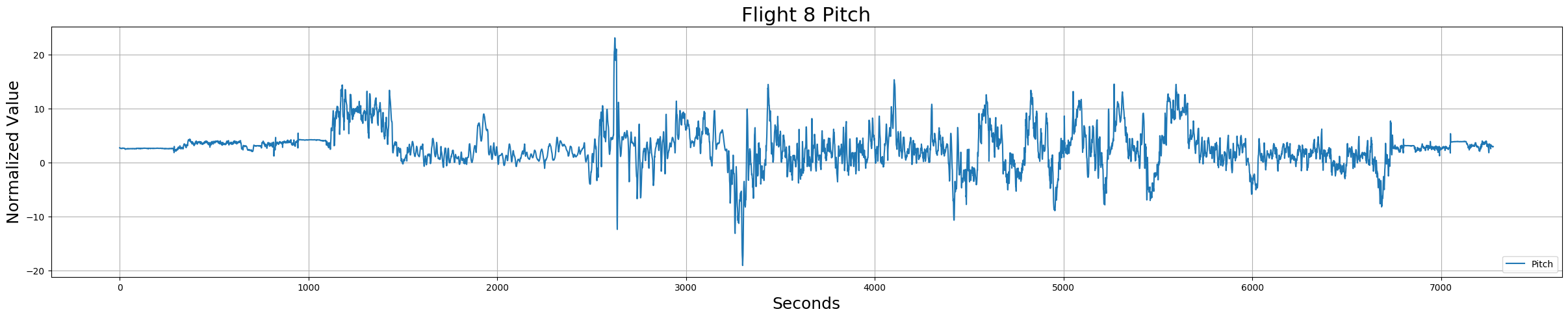}
    \\
    \includegraphics[width=0.98\textwidth]{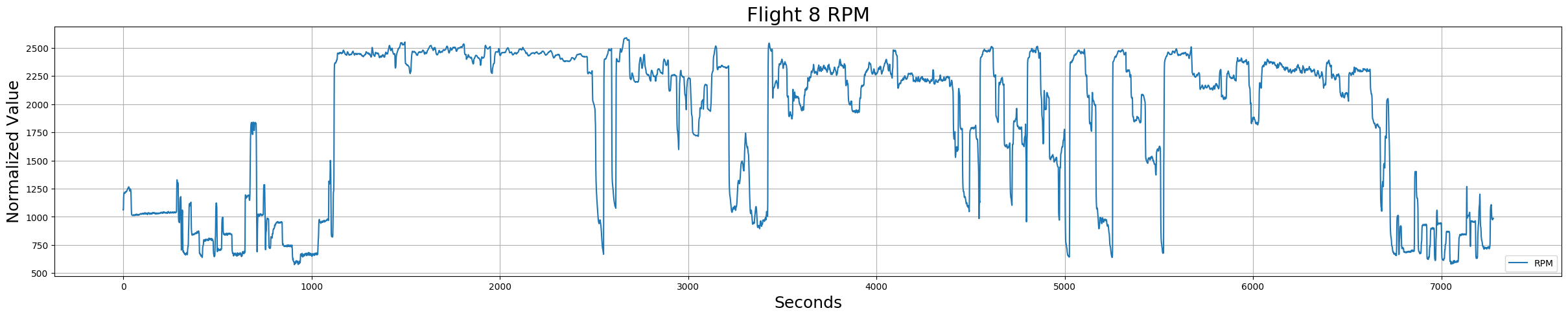}
    \caption{\label{fig:flight_examples} Example parameters pitch (top) and RPM (bottom) of Flight $8$ from the NGAFID dataset.}
\end{figure}

{\it RPM} and {\it pitch} were selected as prediction parameters from the aviation data since RPM is a product of engine activity, with other engine-related parameters being correlated. Pitch itself is directly influenced by pilot controls. As a result, both of these target variables are particularly challenging to predict. 
Figure~\ref{fig:flight_examples} provides an example of the RPM and pitch time series from Flight $8$ of this dataset. In addition, the pitch parameter represents how many degrees above or below horizontal the aircraft is angled. 
As a result, the parameter typically remains steady around a value of $0$, however, it increases or decreases depending on whether or not the aircraft is angled to fly upward or downward, based on pilot controls and external conditions. On the other hand, RPM will mostly vary between an idling speed, i.e., if the plane is on the ground, and a flight speed, with some variation between takeoff and landing. 
Since the majority of the flights in NGAFID (and, by extension, all of the flights in the provided sample) are student training flights, multiple practice takeoffs and landings can be found. This results in two different types of time series, both of which are dependent on the other flight parameters but each with highly different characteristics -- creating excellent time series benchmarks for RNNs.

\subsection{Coal-fired Power Plant Data}
\label{sec:coal_plants}
This dataset consists of $10$ days of per-minute data readings extracted from $12$ out of a coal plant's set of burners. Each of these $12$ data files contains $12$ parameters of time series data: Conditioner Inlet Temp, Conditioner Outlet Temp, Coal Feeder Rate, Primary Air Flow, Primary Air Split, System Secondary Air Flow Total, Secondary Air Flow, Secondary Air Split, Tertiary Air Split, Total Combined Air Flow, Supplementary Fuel Flow, and Main Flame Intensity. 
This data was normalized to the range $[0,1]$, which serves furthermore as a data anonymization step. 

\begin{figure}
    \centering
    \includegraphics[width=0.98\textwidth]{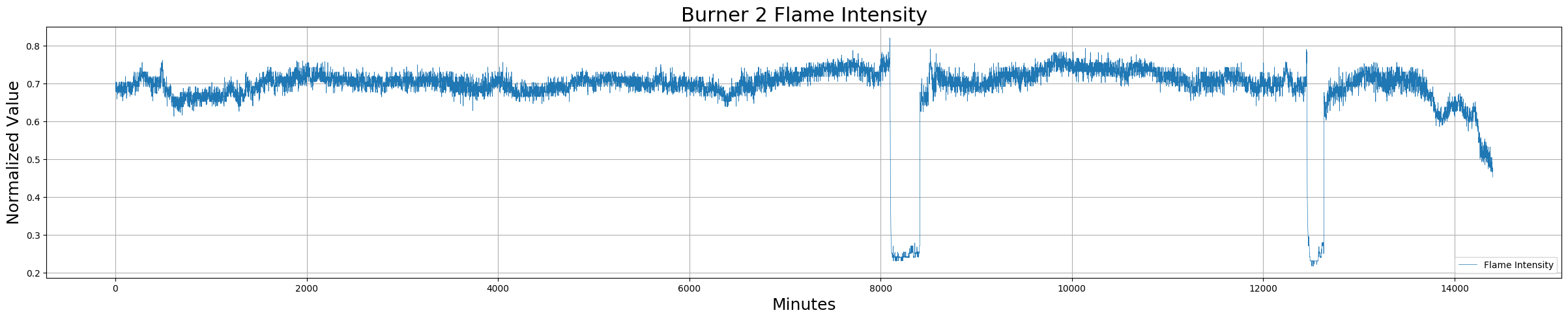}
    \\
    \includegraphics[width=0.98\textwidth]{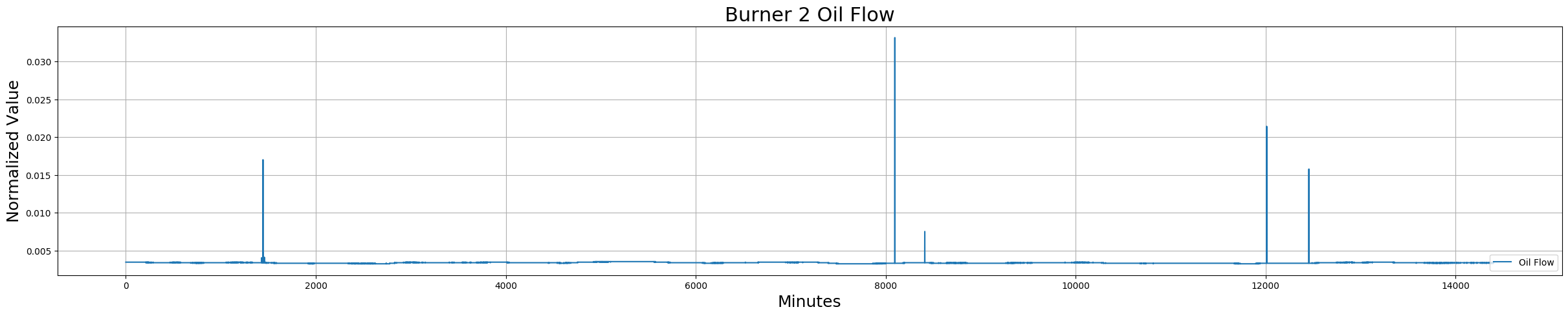}
    \caption{\label{fig:burner_examples} Example parameters for Burner \#2 from the coal plant dataset: flame intensity (top) and fuel flow (bottom).}
\end{figure}

For the coal plant data, \emph{main flame intensity} and \emph{supplementary fuel flow} were selected as parameters of interest. Figure~\ref{fig:burner_examples} provides examples of these two parameters from Burner \# 2 found in the dataset. Main flame intensity is mostly a product of conditions within the burner and parameters related to coal quality which causes it to vary over time. However sometimes planned outages occur or conditions in the burner deteriorate so badly that it is temporarily shut down.  In these cases, sharp spikes occur during the shutdown, which last for an unspecified period of time before the burner turns back on again and the parameter (value) sharply increases. The burners can also potentially operate at different output levels, depending on power generation needs. As a result, step-wise behavior is observed.

On the other hand, supplementary fuel flow remains fairly constant. Nonetheless, it yields sudden and drastic spikes in response to decisions made by plant operators.  When conditions in the burners become poor due to coal quality or other effects, the operator may need to provide supplementary fuel to prevent the burner from going into shutdown.  Of particular interest is if an RNN can successfully detect these spikes given the conditions of the other parameters.  
Similar the key parameters (RPM and pitch) selected in the NGAFID data, main flame intensity is mostly a product of conditions within the (coal) burner while supplementary fuel flow is more directly controlled by human operators. 
Despite these similarities, however, the characteristics of these time series are different from each other as well as from the NGAFID flight data, providing additional, unique benchmark prediction challenges.

%% file: 06-results.tex
\section{Results}
\label{sec:results}

\subsection{Experiments}
\label{sec:experiments}

The first set of ($5$) experiments only permitted the use of a single memory cell type, \ie, exclusively $\Delta$-RNN, GRU, LSTM, MGU, or UGRNN (one experiment per type), and no simple neurons. All of these experiments only allowed the generation of feedforward connections between cells (these experiments were denoted as \emph{delta}, \emph{gru}, \emph{lstm}, \emph{mgu} or \emph{ugrnn}). 
The second set of ($2$) experiments were conducted where the first one only permitted the use of simple neurons and feedforward connections (denoted as \emph{simple}) while the second permitted EXAMM to make use of feedforward connections and simple neurons as well as the choice of any memory cell type (denoted as \emph{all}). 
The next set of experiments ($5$) were identical to the first set with the key exception that EXAMM could choose either between simple neurons and one specified specific memory cell type (these experiments are appended with a \emph{+simple}, i.e., \emph{lstm+simple}). 
The final set of ($12$) experiments consisted of taking the setting of each of the prior $12$ ($5+2+5$) runs and re-ran them but with the modification that EXAMM was permitted to generate deep recurrent connections of varying time delays (these runs are appended with a \emph{+rec}).

This full set of ($24$) experiments was conducted for each of the four prediction parameters, i.e., RPM, pitch, main flame intensity, and supplementary fuel flow. $K$-fold cross validation was carried out for each prediction parameter, with a fold size of $2$. This resulted in $5$ folds for the NGAFID data (as it had $10$ flight data files), and $6$ folds for the coal plant data (as it has $12$ burner data files). 
Each fold and EXAMM experiment was repeated $10$ times.  In total, each of the $24$ EXAMM experiments were conducted $220$ times ($50$ times each for the NGAFID parameter $k$-fold validation and $60$ times each for the coal data parameter $k$-fold validation), for a grand total of $5,280$ separate EXAMM experiments/simulations.

\subsection{EXAMM and Backpropagation Hyperparameters}
\label{sec:examm_bp_hyperparameters}

All RNNs were locally trained with backpropagation through time (BPTT) \cite{werbos1990backpropagation} and stochastic gradient descent (SGD) using the same hyperparameters. SGD was run with a learning rate of $\eta = 0.001$, utilizing Nesterov momentum with $mu = 0.9$. 
No dropout regularization was used since, in prior work, it has been shown to reduce performance when training RNNs for time series prediction~\cite{elsaid2018optimizing}. 
For the LSTM cells that EXAMM could make use of, the forget gate bias had a value of $1.0$ added to it, as \cite{jozefowicz2015empirical} has shown that doing so improves training time significantly. Otherwise, RNN weights were initialized by EXAMM's Lamarckian strategy.

To control for exploding and vanishing gradients, we apply re-scaling to the full gradient of the RNN, $\vec{g}$, which is one single vector of all the partial derivatives of the cost function with respect to the individual weights (in terms of a standard RNN, this amounts to flattening and concatenating all of the individual derivative matrices into one single gradient vector). Re-scaling was done in this way due to the unstructured/unlayered RNNs evolved by EXAMM. Computing the stabilized gradient proceeds formally as follows:
\[
    \vec{g} = 
\begin{cases}
    \vec{g} * \frac{t_h}{\Vert \vec{g}\Vert_2},  & \text{if } \Vert \vec{g}\Vert_2 > t_h \\
    \vec{g} * \frac{t_l}{\Vert \vec{g}\Vert_2},  & \text{if } \Vert \vec{g}\Vert_2 < t_l \\
    \vec{g} & otherwise
\end{cases}
\]
noting that $||\cdot||_2$ is the Euclidean norm operator. $t_h$ is the (high) threshold for preventing diverging gradient values while $t_l$ is the (low) threshold for preventing shrinking gradient values. In essence, the above formula is composed of two types of gradient re-scaling. The first part re-projects the gradient to a unit Gaussian ball (``gradient clipping'' as prescribed by Pascanu \etal~\cite{pascanu2013difficulty}) when the gradient norm exceeds a threshold $t_h = 1.0$. The second part, on the other hand, is a novel trick we introduce called ``gradient boosting'', where, when the norm of the gradient falls below a threshold $t_l = 0.05$, we up-scale it by the factor $\frac{t_l}{||\vec{g}||_2}$.


For EXAMM, each neuro-evolution run consisted of $10$ islands, each with a population size of $5$. New RNNs were generated via intra-island crossover (at a rate of 20\%), mutation at a rate 70\%, and inter-island crossover at 10\% rate. 
All of EXAMM's mutation operations (except for \emph{split edge}) were utilized, each chosen with a uniform 10\% chance.  
The experiments labeled \emph{all} were able to select any type of memory cell or Elman neurons at random,  each with an equal probability. Each EXAMM run generated $2000$ RNNs, with each RNN being trained locally (using the BPTT settings above) for $10$ epochs. Recurrent connections that could span a time skip between $1$ and $10$ could be chosen (uniformly at random). These runs were performed utilizing $20$ processors in parallel, and, on average, required approximately $0.5$ compute hours.  In total, the results we report come from training $10,560,000$ RNNs which required \textasciitilde $52,800$ CPU hours of compute time.

\subsection{Experimental Results}
\label{sec:experimental_results}

\begin{figure*}
    \centering
    \includegraphics[width=0.40\textwidth]{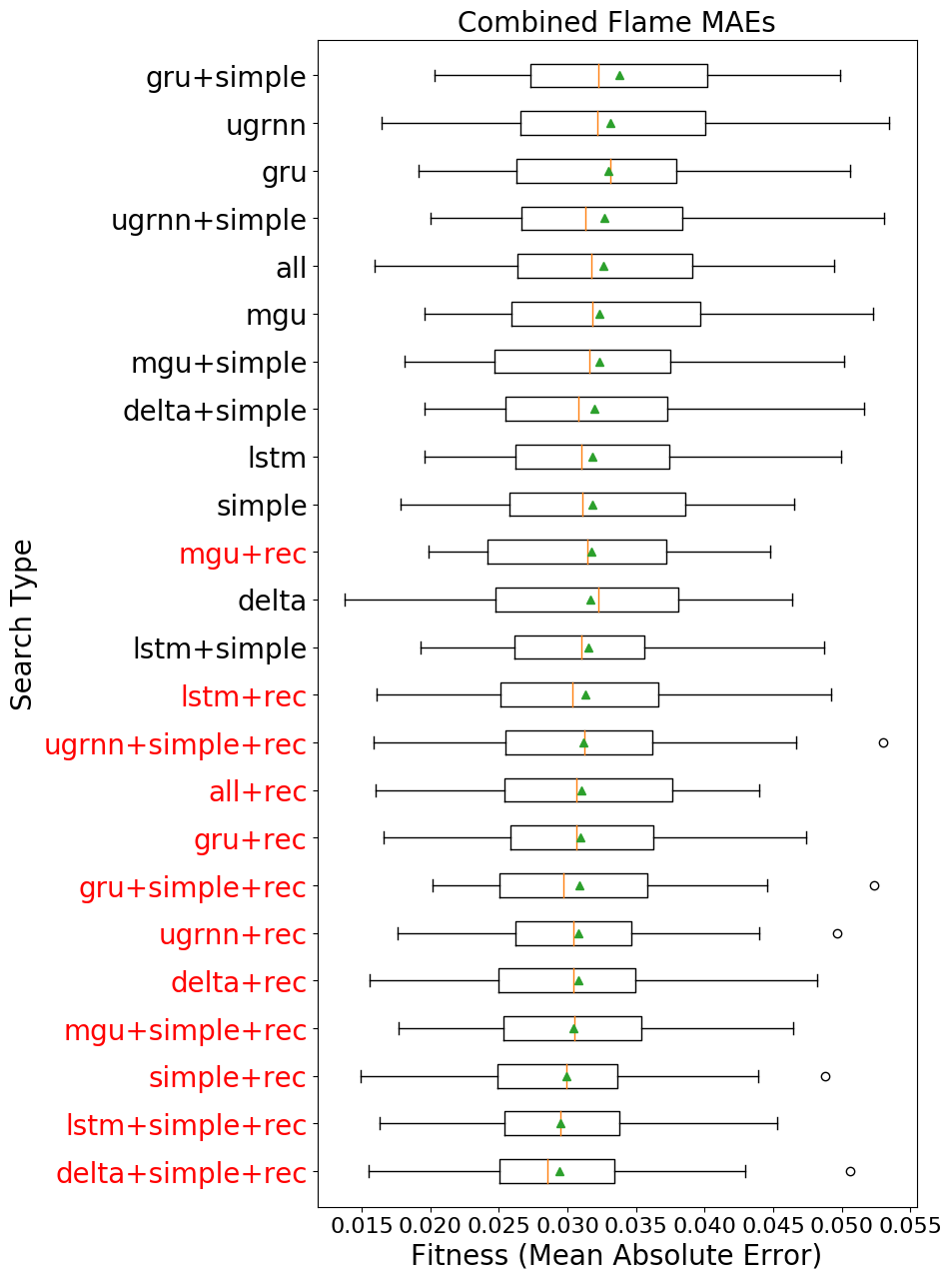}
    \hspace{.5cm}
    \includegraphics[width=0.40\textwidth]{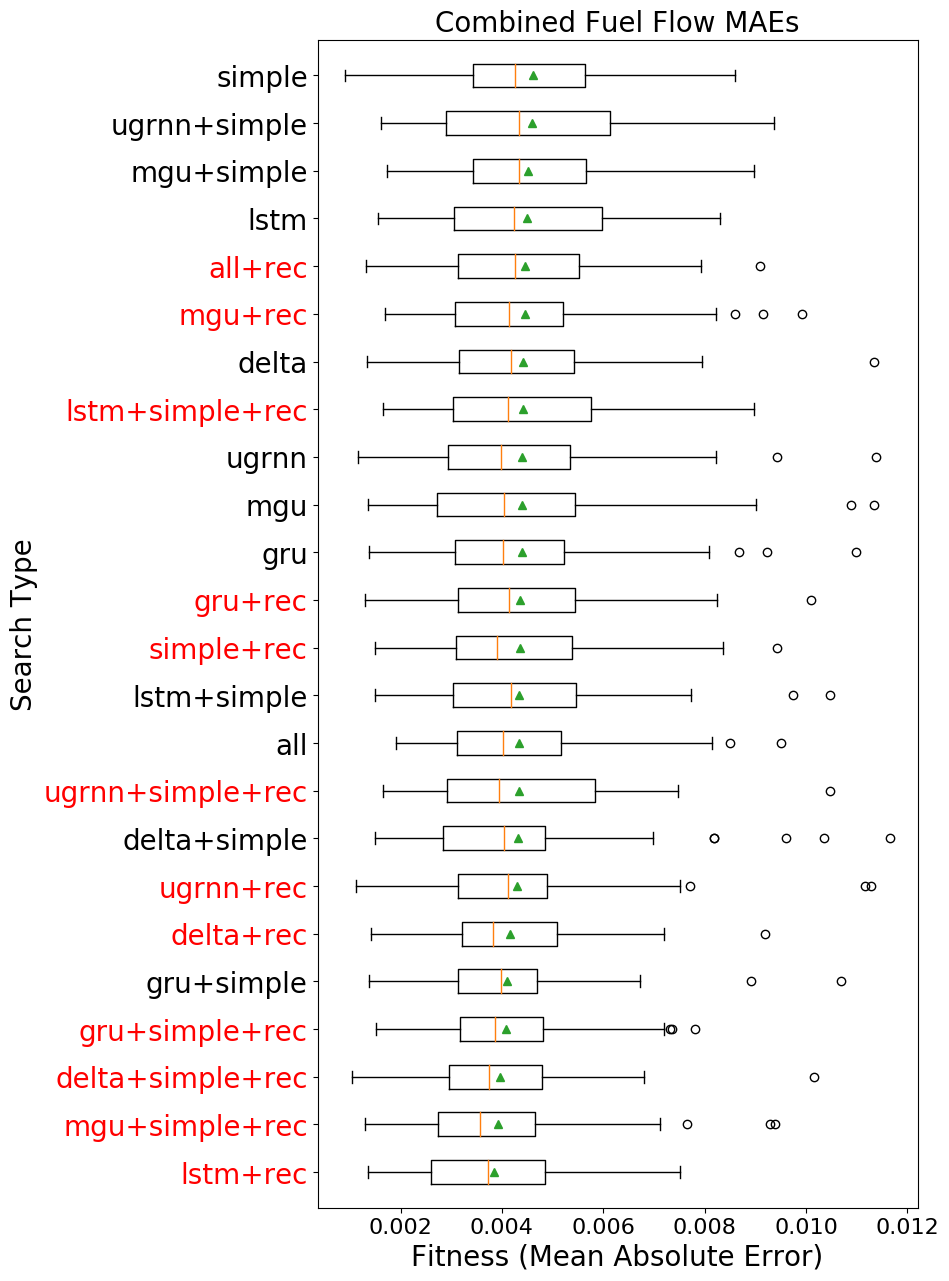}
    \\
    \includegraphics[width=0.40\textwidth]{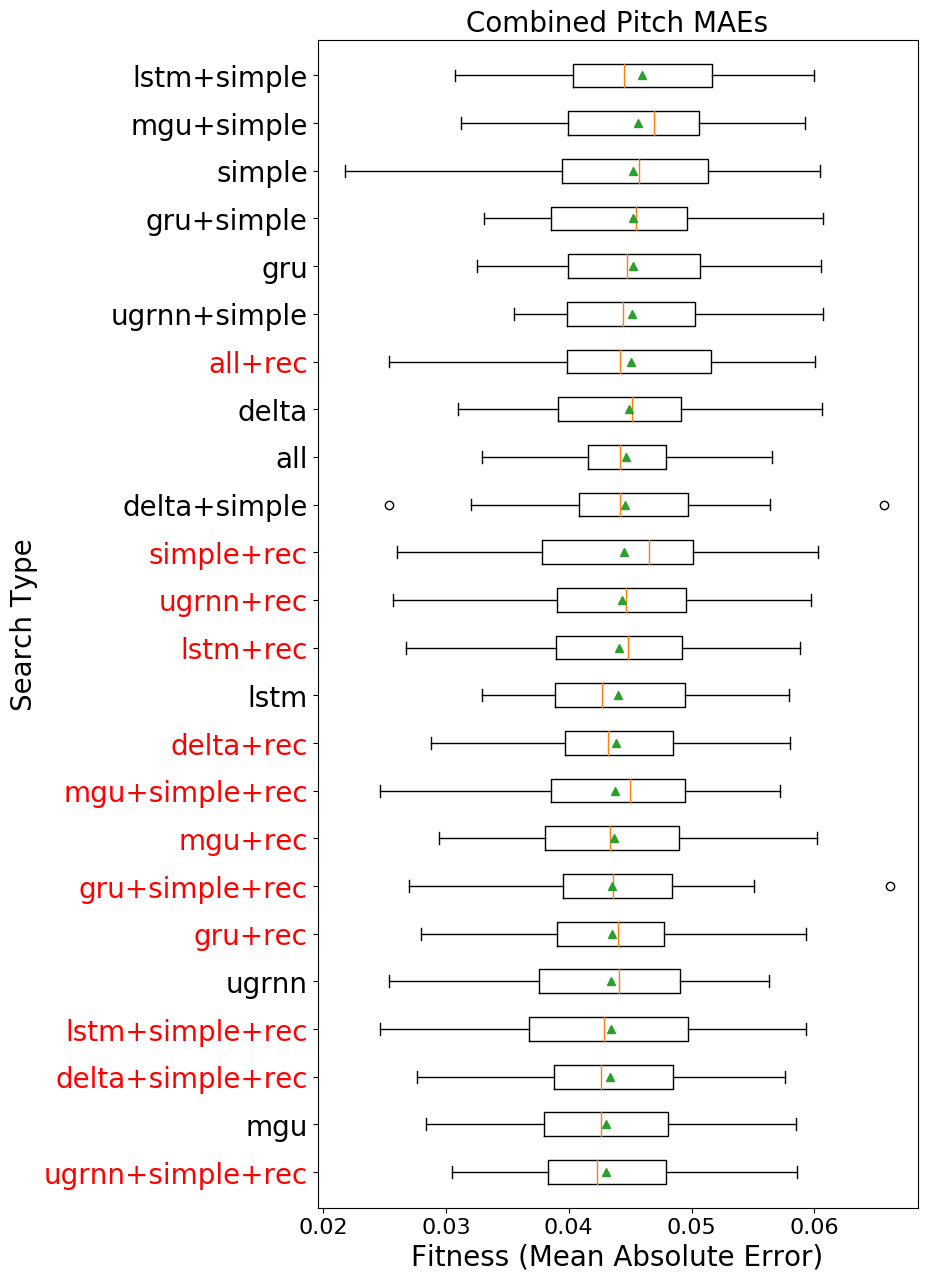}
    \hspace{.5cm}
    \includegraphics[width=0.40\textwidth]{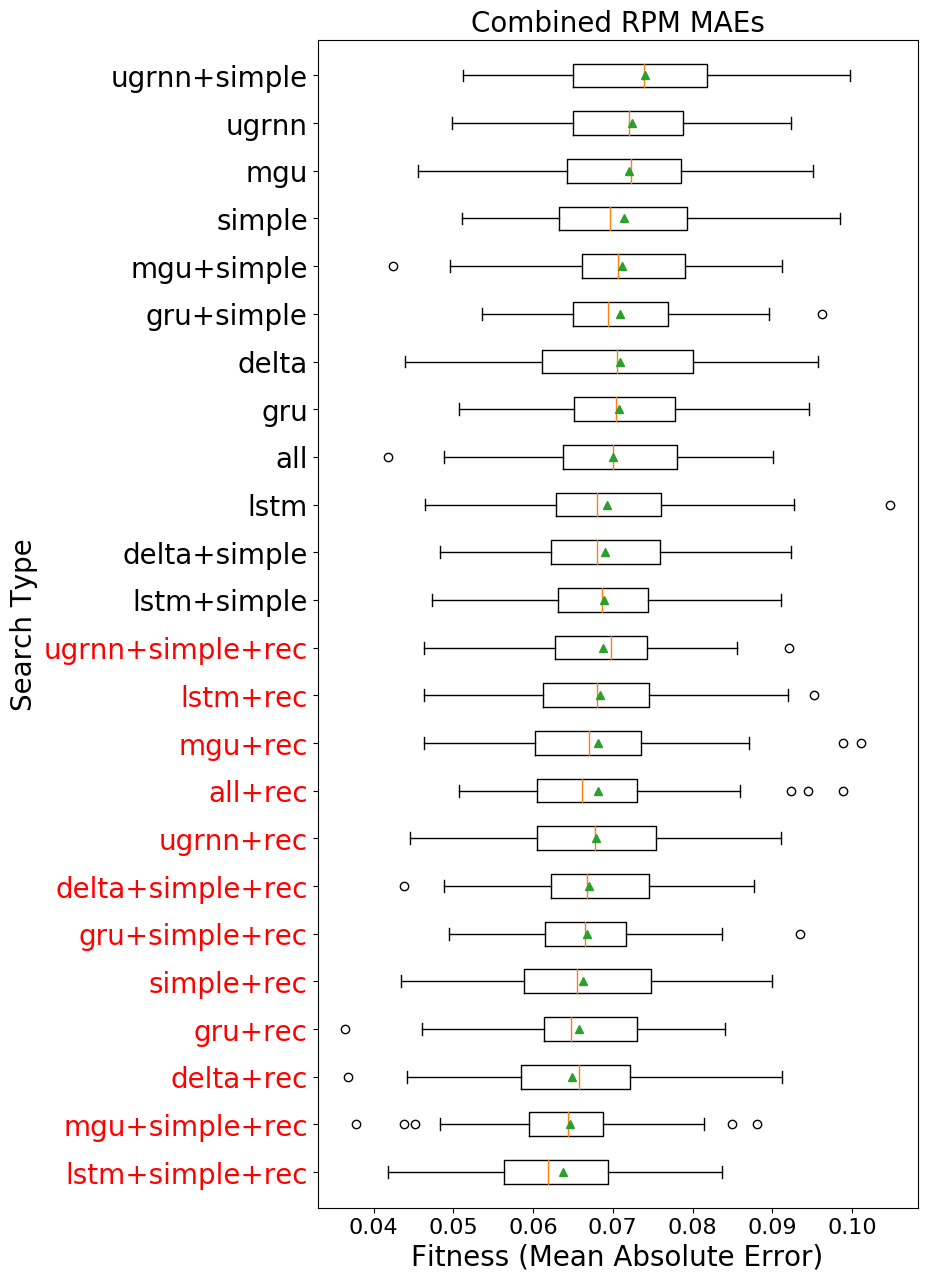}
    \caption{\label{fig:all_fitness} Consolidated range of fitness (mean absolute error) of the best found RNNs for the two datasets' (flame intensity and fuel flow for the coal plant dataset, and pitch and RPM for the aviation datset) target prediction parameters. Results are for the $24$ experiments across all $6$ folds, with $10$ repeats per fold. Run types are ordered top-down by mean.}
\end{figure*}

Figure~\ref{fig:all_fitness} shows the range of the fitness values of the best found neural networks across all of the EXAMM experiments. This combines the results from all folds and all trial repeats -- each box in the box plots represent $110$ different fitness values. The box plots are ordered according to mean fitness (calculated as mean absolute error, or MAE) of the RNNs for that experiment/setting (across all folds), with the top being the highest average MAE, i.e., the worst performing simulation setting, and the bottom containing the lowest average MAE, i.e., the best performing setting. 
Means are represented by green triangles and medians by orange bars. Run type names with deep recurrent connections are highlighted in red.

How well the different experiments performed was also analyzed by calculating the mean and standard deviation of all best evolved fitness scores from each repeated experiment across each fold. This was done since each fold of the test data had a different range of potential best results.  
It was then possible to rank/order the experiments/simulations in terms of their deviation from the mean (providing a less biased metric of improvement). 
Table~\ref{table:avg_devs} presents how well each experiment performed as an average of how many standard deviations they were from the mean in their average case performance. Table~\ref{table:best_devs} is constructed om the same way but this time based on best case performance.
Search types which utilized deep recurrent connections (\emph{+rec}) are highlighted in bold.

\begin{table*}
\begin{subtable}{0.45\textwidth}
\centering
\begin{tabular}{lr}
\hline
Type & Devs from Mean\\
\hline
gru+simple & 0.45466\\
gru & 0.38537\\
mgu+simple & 0.27395\\
ugrnn+simple & 0.26383\\
ugrnn & 0.24459\\
all & 0.22218\\
mgu & 0.21171\\
lstm+simple & 0.05836\\
delta+simple & 0.04819\\
lstm & 0.03707\\
simple & 0.01420\\
delta & 0.00741\\
{\bf ugrnn+simple+rec} & -0.04944\\
{\bf mgu+rec} & -0.05957\\
{\bf lstm+rec} & -0.09531\\
{\bf all+rec} & -0.10439\\
{\bf delta+rec} & -0.13052\\
{\bf gru+rec} & -0.15598\\
{\bf gru+simple+rec} & -0.16988\\
{\bf ugrnn+rec} & -0.18138\\
{\bf mgu+simple+rec} & -0.20289\\
{\bf simple+rec} & -0.30881\\
{\bf lstm+simple+rec} & -0.33808\\
{\bf delta+simple+rec} & -0.42524\\
\hline
\end{tabular}
\caption{$\sigma_{Flame}$: Avg MAE}
\label{table:flame_avg_dev}       
\end{subtable}
\hspace{2cm}
\begin{subtable}{0.45\textwidth}
\centering
\begin{tabular}{lr}
\hline
Type & Devs from Mean\\
\hline
simple & 0.21337\\
mgu+simple & 0.18593\\
ugrnn+simple & 0.17120\\
{\bf all+rec} & 0.12609\\
lstm & 0.11196\\
delta & 0.08607\\
{\bf lstm+simple+rec} & 0.07886\\
{\bf mgu+rec} & 0.07732\\
gru & 0.04823\\
ugrnn & 0.04216\\
{\bf gru+rec} & 0.03564\\
{\bf simple+rec} & 0.01922\\
lstm+simple & 0.01803\\
mgu & 0.00344\\
{\bf ugrnn+simple+rec} & -0.00456\\
all & -0.00649\\
{\bf ugrnn+rec} & -0.02236\\
delta+simple & -0.03849\\
{\bf delta+rec} & -0.08420\\
{\bf gru+simple+rec} & -0.12883\\
gru+simple & -0.13948\\
{\bf delta+simple+rec} & -0.22414\\
{\bf lstm+rec} & -0.28345\\
{\bf mgu+simple+rec} & -0.28552\\
\hline
\end{tabular}
\caption{$\sigma_{Fuel Flow}$: Avg MAE}
\label{table:oil_avg_dev}       
\end{subtable}
\\
\vspace{.5cm}
\begin{subtable}{0.45\textwidth}
\centering
\begin{tabular}{lr}
\hline
Type & Devs from Mean\\
\hline
lstm+simple & 0.23909\\
mgu+simple & 0.20029\\
ugrnn+simple & 0.14481\\
simple & 0.14432\\
gru+simple & 0.13351\\
gru & 0.13152\\
delta & 0.10322\\
{\bf all+rec} & 0.09201\\
all & 0.06990\\
delta+simple & 0.05982\\
{\bf simple+rec} & 0.03136\\
{\bf ugrnn+rec} & 0.00369\\
{\bf lstm+rec} & -0.03673\\
lstm & -0.03996\\
{\bf delta+rec} & -0.06630\\
{\bf mgu+rec} & -0.08360\\
{\bf mgu+simple+rec} & -0.08452\\
{\bf gru+rec} & -0.11290\\
{\bf gru+simple+rec} & -0.11699\\
ugrnn & -0.13281\\
{\bf lstm+simple+rec} & -0.14307\\
{\bf delta+simple+rec} & -0.15425\\
{\bf ugrnn+simple+rec} & -0.18647\\
mgu & -0.19593\\
\hline
\end{tabular}
\caption{$\sigma_{Flame}$: Avg MAE}
\label{table:pitch_avg_dev}       
\end{subtable}
\hspace{2cm}
\begin{subtable}{0.45\textwidth}
\centering
\begin{tabular}{lr}
\hline
Type & Devs from Mean\\
\hline
ugrnn+simple & 0.49582\\
ugrnn & 0.32738\\
mgu & 0.30650\\
simple & 0.26625\\
mgu+simple & 0.23401\\
gru+simple & 0.20177\\
gru & 0.19246\\
delta & 0.17695\\
all & 0.10472\\
lstm & 0.02485\\
lstm+simple & 0.01335\\
delta+simple & 0.00523\\
{\bf ugrnn+simple+rec} & -0.00490\\
{\bf lstm+rec} & -0.02911\\
{\bf mgu+rec} & -0.06871\\
{\bf all+rec} & -0.08791\\
{\bf ugrnn+rec} & -0.09029\\
{\bf delta+simple+rec} & -0.15970\\
{\bf gru+simple+rec} & -0.17044\\
{\bf simple+rec} & -0.21730\\
{\bf gru+rec} & -0.27053\\
{\bf delta+rec} & -0.36819\\
{\bf mgu+simple+rec} & -0.40717\\
{\bf lstm+simple+rec} & -0.47505\\
\hline
\end{tabular}
\caption{$\sigma_{Flame}$: Avg MAE}
\label{table:rpm_avg_dev}       
\end{subtable}
\caption{Average fitness performance values reported for each EXAMM experimental setting. Experimental settings are ranked by their number of standard deviations from the mean of all experiments. Lower values had better performance.}
\label{table:avg_devs}
\end{table*}


\begin{table*}
\begin{subtable}{0.45\textwidth}
\centering
\begin{tabular}{lr}
\hline
Type & Devs from Mean\\
\hline
gru+simple & -1.02844\\
{\bf mgu+rec} & -1.15701\\
ugrnn+simple & -1.21079\\
mgu & -1.24655\\
mgu+simple & -1.26880\\
gru & -1.29390\\
simple & -1.30901\\
lstm+simple & -1.35475\\
lstm & -1.35496\\
delta+simple & -1.37473\\
{\bf ugrnn+rec} & -1.42362\\
ugrnn & -1.43371\\
delta & -1.48912\\
{\bf mgu+simple+rec} & -1.55717\\
{\bf gru+simple+rec} & -1.58618\\
{\bf lstm+simple+rec} & -1.63655\\
{\bf all+rec} & -1.64301\\
all & -1.66893\\
{\bf lstm+rec} & -1.70057\\
{\bf ugrnn+simple+rec} & -1.71172\\
{\bf gru+rec} & -1.73098\\
{\bf delta+rec} & -1.95685\\
{\bf simple+rec} & -1.97756\\
{\bf delta+simple+rec} & -2.08205\\
\hline
\end{tabular}
\caption{$\sigma_{Flame}$: Best MAE}
\label{table:flame_best_dev}       
\end{subtable}
\hspace{2cm}
\begin{subtable}{0.45\textwidth}
\centering
\begin{tabular}{lr}
\hline
Type & Devs from Mean\\
\hline
{\bf gru+rec} & -1.10116\\
ugrnn+simple & -1.18567\\
lstm+simple & -1.18625\\
{\bf mgu+rec} & -1.18778\\
{\bf lstm+simple+rec} & -1.21500\\
mgu+simple & -1.21509\\
all & -1.22138\\
gru+simple & -1.27796\\
{\bf gru+simple+rec} & -1.29070\\
{\bf simple+rec} & -1.29699\\
simple & -1.30479\\
ugrnn & -1.30559\\
{\bf ugrnn+simple+rec} & -1.31366\\
delta & -1.33034\\
{\bf delta+rec} & -1.35481\\
{\bf all+rec} & -1.37338\\
lstm & -1.38003\\
delta+simple & -1.38368\\
{\bf lstm+rec} & -1.38510\\
{\bf ugrnn+rec} & -1.42369\\
mgu & -1.45259\\
{\bf mgu+simple+rec} & -1.50962\\
gru & -1.53812\\
{\bf delta+simple+rec} & -1.54667\\
\hline
\end{tabular}
\caption{$\sigma_{Fuel Flow}$: Best MAE}
\label{table:oil_best_dev}       
\end{subtable}
\vspace{0.5cm}
\\
\begin{subtable}{0.45\textwidth}
\centering
\begin{tabular}{lr}
\hline
Type & Devs from Mean\\
\hline
ugrnn+simple & -0.99073\\
gru & -1.01889\\
lstm+simple & -1.09707\\
gru+simple & -1.10143\\
delta & -1.19651\\
lstm & -1.24966\\
all & -1.25872\\
{\bf delta+rec} & -1.42943\\
mgu+simple & -1.48976\\
{\bf all+rec} & -1.55755\\
{\bf ugrnn+rec} & -1.58235\\
{\bf mgu+rec} & -1.60397\\
{\bf lstm+rec} & -1.63888\\
{\bf ugrnn+simple+rec} & -1.64192\\
mgu & -1.67690\\
ugrnn & -1.70299\\
{\bf delta+simple+rec} & -1.77567\\
delta+simple & -1.78042\\
{\bf gru+rec} & -1.81352\\
{\bf lstm+simple+rec} & -1.89858\\
simple & -2.05128\\
{\bf mgu+simple+rec} & -2.09451\\
{\bf gru+simple+rec} & -2.09545\\
{\bf simple+rec} & -2.24764\\
\hline
\end{tabular}
\caption{$\sigma_{Pitch}$: Best MAE}
\label{table:pitch_best_dev}       
\end{subtable}
\hspace{2cm}
\begin{subtable}{0.45\textwidth}
\centering
\begin{tabular}{lr}
\hline
Type & Devs from Mean\\
\hline
gru & -0.94516\\
simple & -0.99991\\
gru+simple & -1.08121\\
mgu & -1.17371\\
ugrnn+simple & -1.19714\\
{\bf all+rec} & -1.34347\\
ugrnn & -1.36917\\
{\bf ugrnn+simple+rec} & -1.44366\\
{\bf gru+simple+rec} & -1.49508\\
mgu+simple & -1.49991\\
lstm & -1.50167\\
{\bf delta+simple+rec} & -1.51271\\
delta+simple & -1.51795\\
{\bf mgu+rec} & -1.52494\\
delta & -1.57259\\
lstm+simple & -1.64965\\
all & -1.69526\\
{\bf lstm+simple+rec} & -1.71450\\
{\bf ugrnn+rec} & -1.72680\\
{\bf lstm+rec} & -1.74024\\
{\bf simple+rec} & -1.74335\\
{\bf gru+rec} & -1.88070\\
{\bf mgu+simple+rec} & -1.89718\\
{\bf delta+rec} & -2.05063\\
\hline
\end{tabular}
\caption{$\sigma_{RPM}$: Best MAE}
\label{table:rpm_best_dev}       
\end{subtable}
\caption{Best fitness performance values reported for each EXAMM experimental setting. Experimental settings are ranked by their number of standard deviations from the mean of all experiments. Lower values had better performance.}
\label{table:best_devs}
\end{table*}

\subsection{Memory Cell Performance}
\label{sec:mem_cell_performance}

\begin{table}
\centering
\begin{tabular}{lrrrr}
\hline
    Memory  & \multicolumn{2}{c}{Top 3} & \multicolumn{2}{c}{Bottom 3} \\
Cell    & Avg   & Best  & Avg       & Best\\
\hline
    all     & 0 & 0 & 0 & 0 \\
    simple  & 1 & 2 & 1 & 1 \\
    delta   & 4 & 4 & 0 & 0 \\
    gru     & 0 & 3 & 2 & 5 \\
    lstm    & 3 & 0 & 1 & 2 \\
    mgu     & 2 & 3 & 4 & 1 \\
    ugrnn   & 1 & 0 & 4 & 3 \\
\hline
\end{tabular}
\caption{How often a memory cell type appeared in the top $3$ or bottom $3$ experiments in the best and average cases.}
\label{table:top_bottom_counts}       
\end{table}

Table~\ref{table:top_bottom_counts} shows the frequency of a particular memory cell experiment/setting appearing in the three best (Top 3) or three worst (Bottom 3) slots (in a ranked list) for each prediction parameter for the experiment's average and best RNN fitness score. 
The \emph{simple} row includes only the \emph{simple} and \emph{simple+rec} runs and the \emph{all} row includes the \emph{all} and \emph{all+rec} runs, while the other memory cell rows include the \emph{+simple} and \emph{+rec} versions (\eg, the \emph{delta} row includes occurrences of \emph{delta}, \emph{delta+simple}, \emph{delta+rec} and \emph{delta+simple+rec}).

Based on these count results, the $\Delta$-RNN memory cells performed the best, appearing in the top 3 for the average and best cases $4$ times each. Furthermore, it did not appear in the bottom $3$ for the average and best cases \emph{at all}, making these particular memory cells more reliable than the others. MGU memory cells performed the next best, appearing twice in the average case and $3$ times in the best case. However, they also showed up $4$ times in the bottom, $3$ on the average case, and once in the bottom $3$ for the best case networks. Interestingly enough, while the popular LSTM memory cells showed up frequently for the average performance case ($3$ times), they did not show up at all in the top $3$ for the best found RNNs. They also occurred once in the bottom three for average performance and twice in the bottom $3$ for best performance.

The experimental results indicate simple neurons performed rather well when combined with deep recurrent connections. The \emph{simple+rec} configuration showed up once in the top $3$ for the average case, and twice in the top $3$ for the best case. 
When simple neurons appeared in the bottom $3$, it was only in the experiments when no deep recurrent connections were permitted. As a result, aside from the $\Delta$ and MGU memory cells, \emph{simple+rec} performed better than the other more complicated memory cells, e.g., LSTM, GRU, and was, furthermore, more reliable than the MGU cells.

The performance of the GRU memory cells was intriguing -- they showed up $3$ times in the top $3$ for the best case RNNs, $0$ times for the top $3$ average case runs, but $2$ and $5$ times in the bottom $3$ for average and best case networks.  This seems to indicate that, while GRU memory cells have the potential to find well performing networks, they are \emph{highly unreliable} for these datasets. We hypothesize that this might due to either high sensitivity to initialization conditions or to unknown limitations in the way they gate/carry temporal information.

Lastly, UGRNN memory cells performed the worst overall.  They only appeared once in the top $3$ average case and not at all in the top $3$ best case. At the same time they occurred $4$ and $3$ times in the bottom $3$ for the average and best case performance rankings.

The \emph{all} configurations did not show up at all in the top $3$ or bottom $3$, most likely due to the significant size of its particular search space. Given the additional option to select from a large pool of different memory cell types, the EXAMM neuro-evolution procedure might just simply require far more time to decide on optimal cell types that would yield better/top performing networks. 

\subsection{Effects of Simple Neurons}
\label{sec:simple_neurons}

Table~\ref{table:simple_changes} provides measurements for how the addition of simple neurons changed the performance of the varying memory cell types. In it, we show how many standard deviations from the mean the average case moved when averaging the differences of \emph{mgu} to \emph{mgu+simple} and \emph{mgu+rec} to \emph{mgu+simple+rec} (over all four predicition parameters). In the average case, adding simple neurons did appear to yield a modest improvement, improving deviations from the mean by $-0.02$ overall, and improving deviation from the mean for all memory cells except for UGRNNs. 
Adding simple neurons had a similar overall improvement for the best found RNNs, however, this incurred a much wider variance. Despite this variance, $2$ of the $3$ best found networks had \emph{+simple} as an option, with a third being \emph{simple+rec}. This seems to indicate that most memory cell types could either benefit by mixing/combining them with simple neurons.

\subsection{Effects of Deep Recurrent Connections}
\label{sec:deep_reccurence}

Table~\ref{table:recurrent_changes} provides similar measurements for EXAMM settings that permitted the addition of deep recurrent edges to the varying memory cell types, as well as the \emph{all} and \emph{simple} runs. Compared to adding \emph{+simple}, the \emph{+rec} setting showed an order of magnitude difference, improving deviations from the mean by $-0.2$ overall. In addition, for each of the prediction parameters, the best found RNN utilized deep recurent connections. Looking at the top $3$ best and top $3$ average case RNNs, $11$ out of $12$ utilized deep recurrent connections. 
Similarly, in the bottom $3$ best, \emph{+rec} occurs twice and does not appear at all in the bottom $3$ average case run types. For the Flame and RPM parameters, on the average case, even the worst performing run type with \emph{+rec} performs better than any experiments without it.

\begin{table}
\centering
\begin{tabular}{lrr}
\hline
Type & Dev for Avg & Dev for Best \\
\hline
delta & -0.07663 & -0.07420 \\
gru & -0.02369 & 0.04575 \\
lstm & -0.02973 & 0.02485 \\
mgu & -0.03463 & -0.18857 \\
ugrnn & 0.07991 & 0.15908 \\
overall & -0.02365 & -0.02018 \\
\hline
\end{tabular}
\caption{Performance improvement (in std. devs from the mean) for adding simple neurons.}
\label{table:simple_changes}       
\end{table}

\begin{table}
\centering
\begin{tabular}{lrr}
\hline
Type & Dev for Avg & Dev for Best \\
\hline
all & -0.09113 & -0.01828 \\
simple & -0.27842 & -0.40014 \\
delta & -0.25571 & -0.30079 \\
gru & -0.31534 & -0.43257 \\
lstm & -0.14463 & -0.24462 \\
mgu & -0.11507 & 0.01901 \\
ugrnn & -0.19291 & -0.08625 \\
overall & -0.19903 & -0.20909 \\
\hline
\end{tabular}
\caption{Performance Improvement (in std. devs from the mean) for adding deep recurrent connections.}
\label{table:recurrent_changes}       
\end{table}

%% file: 07-discussion.tex
\section{Discussion}
\label{sec:discussion}

The results presented in this work contribute some significant and interesting insights for RNN-based time series data prediction. The main findings of this study are as follows:

\begin{itemize}
\item \emph{Deep Recurrent Connections:} yielded the most significant improvements in RNN generalization, and, in some cases, were more important than the use of memory cells, i.e., the \emph{simple+rec} experiments performing quite strongly. For all four benchmark datasets, the best found RNNs included those that made use of deep recurrence. As a whole, adding deep recurrent connections to the evolutionary process resulted in large shifts of improvement in the standard deviations from mean measurement. 
These results are particularly significant given that the commonly accepted story is that one should primarily use LSTM or other gated neural structures in order to stand a chance at capturing long term time dependencies in temporal data (despite the fact that internal connections only explicitly traverse a single time step) when classically it has been known that time delays and temporal skip connections can vastly improve generalization over sequences.

\item \emph{Strong simple+rec Performance:} Another very interesting finding was that only using simple neurons and deep recurrent connections, without any memory cells, (the \emph{simple+rec} experiment) performed quite well. This found the best RNN with respect to the Pitch prediction problem (aviation), the second best on the Flame prediction problem dataset (coal), and the fourth best on the RPM prediction problem (aviation).  This shows that, in some cases, it may be more important to have deep recurrent connections than more complicated memory cells.  
    
\item \emph{Strong $\Delta$-RNN Memory Cell Performance:} While there is no ``free lunch''  in statistical learning or optimization \cite{wolpert1997no}, the newer $\Delta$-RNN memory cell did consistently stand out as one of the better-performing memory cells  In three out of the four datasets, EXAMM found it to be the best performing RNN cell-of-choice, and for the average case performance, the $\Delta$-RNN made it into the top $3$ experiments for all four datasets.  
Furthermore, unlike the other memory cell experiments $\Delta$-RNN did not appear in the bottom $3$ \emph{for any of the experiments, either in the average or best cases}.  The only other experiment setting/configuration to boast top $3$ best performance and no bottom $3$ performance was the \emph{simple+rec} experiment. However this did not perform as well in the average case, only appearing in the top $3$ twice. Our results showing that the $\Delta$-RNN consistently outperforms more complex cells such as the LSTM corroborates the findings of \cite{ororbia2017diff}, which presented early findings in the domain of language modeling.  While a newer memory cell, our results indicate that, while deep recurrence and time delay are critical for robustly modeling sequences, simpler gated cells like the $\Delta$-RNN cell should also be strongly considered when designing RNNs, especially for time series forecasting.

\end{itemize}

\section{Future Work}
\label{sec:future}

The choice of selecting time skip depths uniformly at random between the hyperparameter range $[1,10]$ was a somewhat arbitrary choice. We hypothesize that an adaptive approach to selecting the depth skip (or length of the time delay) based on previously well-performing configurations/model candidates might provide better accuracy and remove the need for choosing the bounds of time delay range.  
Perhaps the most interesting direction to pursue is to develop memory cells that efficiently and effectively use recurrent connections that explicitly span more than one step in time, i.e., perhaps more intelligent/powerful gating mechanisms could be design to properly mix together the information that flows from multiple time delays. In addition, perhaps EXAMM can be used to aumatically incorporate or design better variations of highway connections as well, given the potential expressive power that recurrent highway networks \cite{zilly2017recurrent} offer.


The strong performance of the \emph{simple+rec} experiment might also suggest that generating and training RNNs using an evolutionary process with Lamarckian weight initialization may make training RNNs with non-gated recurrent connections easier. This naturally happens since neuro-evolution process such as EXAMM will discard poor RNN solutions that occur in the search space, i.e., poor minima/regions that result from exploding or vanishing gradients when using backpropagation through time (BPTT), and not add them to its candidate solution population, preventing the generation of at least offspring that generalize too poorly. As a result, the evolutionary process will tend to preserve RNNs which have been training well (or at least, when trained with BPTT, have well-behaved gradients). Future investigation can explore if this is truly the case by by retraining the best found architectures from scratch and comparing their performance across various sequence modeling settings.



\section{Conclusion}
\label{sec:conclusion}

While most work in the field of neuro-evolution focuses on the evolution of neural architectures that can potentially outperform hand-crafted designs, this work showcases the potential of neuro-evolution for a different use:  a robust analysis and investigation of the performance and capabilities of different artificial neural network components. Specifically, we demonstrate EXAMM as powerful tool for analyzing/designing recurrent networks, focused on the choice of internal memory cells and the density and complexity of recurrent connectivity patterns. Rigorously investigating a new neural processing component can be quite challenging given that, often, its performance is tied to the overall architecture it is used within. For most work, new architectural components or strategies are typically only investigated within a few select architectures which may not necessarily represent how well the processing mechanism would perform given a much wider range of potential architectures it could be integrated into. 
Neuro-evolution helps alleviate this problem by allowing the the structural components themselves to play a key role in determining the architecture/systems they will most likely work well within. This facilitates a far more fair comparison of their capabilities and, perhaps, allows us to draw more general insights in our quest to construct robust neural models that generalize well.